# Knowledge Graph-Infused Fine-Tuning for Structured Reasoning in Large Language Models

Wuyang Zhang*
University of Massachusetts Amherst
Amherst, USA

Yexin Tian
Georgia Institute of Technology
Atlanta, USA

Xiandong Meng
University of California, Davis
Davis, USA

Mengjie Wang
New York University
New York, USA

Junliang Du
Shanghai Jiao Tong University
Shanghai, China

*Abstract*-This paper addresses the problems of missing reasoning chains and insufficient entity-level semantic understanding in large language models when dealing with tasks that require structured knowledge. It proposes a fine-tuning algorithm framework based on knowledge graph injection. The method builds on pretrained language models and introduces structured graph information for auxiliary learning. A graph neural network is used to encode entities and their relations, constructing a graph-based semantic representation. A fusion mechanism is then designed to jointly model the knowledge graph embeddings with the contextual representations from the language model. To enhance the robustness of knowledge integration, a gating mechanism is introduced to dynamically balance the contributions of linguistic semantics and structural knowledge. This effectively mitigates conflicts between different representational spaces. During training, a joint loss function is constructed to account for both task performance and structural alignment objectives. This helps improve the accuracy of entity prediction and semantic reasoning. The study also includes a series of systematic sensitivity experiments. It evaluates the effects of learning rate, graph coverage, and structural perturbations on model performance. The results further validate the effectiveness and stability of the proposed method across tasks such as entity recognition, question answering, and language generation. Experimental findings show that the proposed structure-aware fine-tuning framework significantly enhances the model's ability to represent complex semantic units. It demonstrates better semantic consistency and contextual logic modeling in scenarios involving structural reasoning and entity extraction.

## I. INTRODUCTION

With the rapid advancement of artificial intelligence, large language models (LLMs) have emerged as a core technology in natural language processing (NLP). These models, powered by massive datasets and complex parameter structures, demonstrate remarkable capabilities in language understanding, generation, and reasoning. However, despite their success in general-purpose tasks, LLMs still face challenges in domain-specific applications that require structured knowledge. Issues such as sparse knowledge, incomplete reasoning chains, and persistent semantic ambiguities remain unresolved[1]. These limitations largely stem from the inadequate use of explicit knowledge during model training, especially the lack of deep injection mechanisms for structured semantic information. Therefore, injecting domain knowledge into LLMs in a structured manner has become a key approach to enhancing their performance in specific application scenarios.

Knowledge graphs (KGs), as high-level semantic network structures, organize abstract knowledge through entities and relations. They represent complex semantic associations using nodes and edges. Compared to traditional text-based knowledge representation, KGs provide stronger structural organization and computability. They effectively encode hierarchical concepts, attribute constraints, and subsumption relations. Integrating KGs into language models can significantly improve the models' entity recognition and contextual understanding. KGs also provide explicit logical paths that enhance the interpretability and accuracy of reasoning processes. Thus, combining KGs with LLMs is a promising direction for advancing from general semantic modeling to deep domain-specific understanding[2].

Nevertheless, there is an inherent heterogeneity between knowledge graphs and LLMs in terms of structure and training paradigms. LLMs rely on sequential language inputs, while KGs use non-Euclidean graph-based semantic structures. Achieving deep integration of these two forms of representation without disrupting the original architecture of LLMs remains a major research challenge. Furthermore, knowledge graphs themselves suffer from data incompleteness, entity sparsity, and structural diversity, which increase the complexity of knowledge injection. A successful injection mechanism must balance knowledge selection, encoding, and fusion strategies. It must also align the capabilities of LLMs with the complementary strengths of knowledge graphs[3,4].

In real-world applications, many high-risk and knowledge-intensive domains, such as financial analysis [5-6], medical reasoning [7-8], elastic cloud resource scaling [9-10], demand greater expertise and reasoning from language models. In these domains, models must understand complex terminology and infer causal, hierarchical, and event-based relationships between entities. Such knowledge is often unavailable in unstructured texts and must be supplemented with structured sources[11,12]. Therefore, developing a fine-tuning algorithm that dynamically guides the model's cognitive path based on KG structures and maintains semantic consistency and knowledge alignment during text generation has both practical and theoretical importance. This approach enhances model robustness and generalization in specific tasks. It also bridges the gap between knowledge representation and reasoning capabilities[13].

Moreover, as AI systems move toward being explainable, reliable, and domain-proficient, they face issues such as strong black-box characteristics and weak knowledge controllability. Traditional fine-tuning methods rely heavily on data and cannot explicitly use external knowledge. This makes it difficult to meet the safety and verifiability requirements of decision-critical scenarios[14]. Using knowledge graphs as an intermediary to guide information structures enables semantic generation and knowledge reasoning to be grounded in verifiable fact networks. This improves transparency and controllability in decision-making. Such integration complements the conventional pretraining – fine-tuning paradigm. It also provides a practical foundation and cutting-edge direction for deploying LLMs in knowledge-intensive tasks.

## II. PROPOSED APPROACH

This study presents a fine-tuning algorithm for large language models, centered on knowledge graph injection, with the goal of strengthening both representation and reasoning capacities by leveraging structured knowledge. The overall architecture draws inspiration from Lyu et al.[15], who demonstrated that modular modeling strategies—combining a pre-trained language model with specialized guidance modules—can significantly improve performance on knowledge-intensive tasks. In this framework, the architecture comprises two primary components: the pre-trained language model, which captures contextual semantic representations, and a knowledge graph guidance module designed to complement semantic units with graph structure and facilitate relational reasoning. During training, consistent with the methodology outlined by Wang, knowledge graph information is first encoded into graph embeddings, enabling a structured representation of entities and relations. These embeddings are then fused with the latent vectors within the language model, dynamically guiding semantic learning and enhancing robustness against structural noise and adversarial input [16]. To further enable complex reasoning and support context-aware understanding, this work adopts a fusion-based approach similar to that proposed by Sun et al., where multi-source knowledge representations are integrated at the semantic level for more effective context–knowledge alignment [17]. The synergy between the language model and the knowledge graph guidance module is illustrated in Figure 1, providing a basis for structured semantic learning and knowledge-augmented inference within the proposed framework.

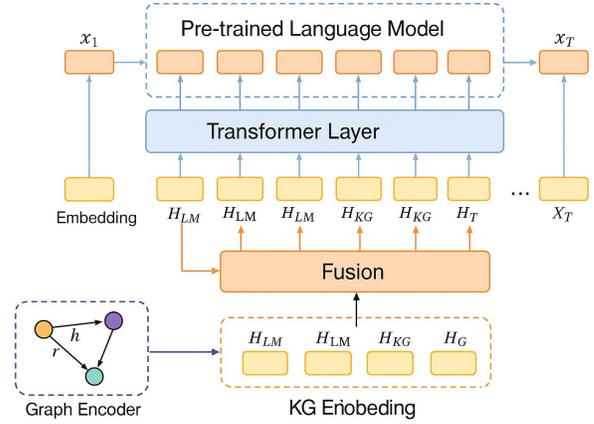

Figure 1. Overall model architecture diagram

In the knowledge encoding part, the triple $(h, r, t)$ is first mapped to a low-dimensional dense vector, and the knowledge graph is encoded through a graph neural network (such as GCN or R-GCN), which is formalized as follows:

$$h_v^{(l+1)} = \sigma \left( \sum_{u \in N(v)} \frac{1}{c_{uv}} W_r^{(l)} h_u^{(l)} \right) \quad (1)$$

Among them, $h_v^{(l)}$ represents the representation of node v in layer l, $N(v)$ is its neighbor node set, $c_{uv}$ is the normalization factor, $W_r^{(l)}$ is the weight matrix related to edge type $r$, and $\sigma$ is the activation function. After multiple layers of graph convolution, the obtained knowledge representation will be injected into the language model as a structural prior.

To optimize the integration of contextual semantics and structured knowledge, this framework incorporates a gating unit that dynamically regulates the fusion ratio between the language model's contextual representations and the knowledge graph-derived information. The design of this gating mechanism is influenced by the attention-based fusion strategies proposed by Xu et al., who demonstrated that adaptive gating can significantly improve the selective integration of heterogeneous information sources in complex clinical NLP scenarios [18]. By enabling the model to assign variable weights to different knowledge streams, the gating unit ensures that relevant knowledge is emphasized for each specific context. Additionally, Meng et al. provided insights into robust information fusion in distributed and federated learning settings, where dynamic gating is crucial for handling multi-source data and enhancing anomaly detection performance [19]. Drawing on Zhu's recent advances in rapid adaptation for language models, the gating unit within this framework also supports efficient gradient-based adjustment, allowing the model to swiftly respond to changes in the structure or content of incoming knowledge graphs [20]. This dynamic gating approach helps the model achieve a more precise and context-sensitive balance between linguistic and structural knowledge, ultimately supporting more robust and explainable knowledge integration. Specifically, the context representation generated by the language model is defined as

$H_{LM}$, and the knowledge graph encoding is defined as $H_{KG}$. The fused representation is:

$$H_{fused} = \lambda \cdot H_{LM} + (1-\lambda) \cdot H_{KG} \quad (2)$$

Among them, $\lambda = \sigma(W_g[H_{LM}; H_{KG}] + b_g)$ is the gating factor controlled by learnable parameters, $[\cdot;\cdot]$ which represents the vector concatenation operation, and $\sigma$ is the Sigmoid function, which is used to ensure that the fusion weight is within the $[0;1]$ interval.

In the fine-tuning stage, a joint loss function is applied to ensure the fused representations are well-optimized for downstream tasks. This loss function combines the main task loss—such as language modeling loss—with an additional term designed to align the model's outputs with the structured knowledge provided by the graph. Qin's work on hierarchical semantic-structural encoding demonstrates the effectiveness of using such joint objectives, as this approach helps the model maintain both semantic consistency and accurate structural alignment throughout the learning process [21]. To make the alignment loss more effective for different task scenarios, this study incorporates principles from Wu, who emphasizes task-aware structural regularization to facilitate parameter-efficient fine-tuning and to adapt flexibly to diverse downstream requirements [22]. Moreover, drawing on the distributed training strategies highlighted by Zhang et al., the joint loss also supports robust and stable optimization, even when dealing with data from varied sources or domains [23]. This approach allows the model to preserve both linguistic accuracy and structural integrity, as formalized below:

$$L_{total} = L_{task} + \alpha \cdot L_{align} \quad (4)$$

Among them, $L_{task}$ represents the loss of tasks such as language modeling or question answering $L_{align} = \sum_i \| H_{LM}^{(i)} - H_{KG}^{(i)} \|^2$ represents the Euclidean distance alignment term between language representation and knowledge representation, and $\alpha$ is a hyperparameter that controls the balance of the two types of information during training.

To further improve the model's ability to model graph structure relationships, a knowledge-aware attention mechanism is constructed to emphasize the semantic dependencies between entities. For the i-th word in the context, its final representation is:

$$z_i = \sum_{j \in \varepsilon} \text{softmax}_j \left( \frac{(Qh_i)^T (Kh_j)}{\sqrt{d}} \right) \cdot Vh_j \quad (5)$$

Among them, $\varepsilon$ is the entity set related to word i, $Q, K, V$ is the linear transformation matrix of query, key, and value, respectively, and d is the scaling factor. This mechanism explicitly introduces entity nodes in the graph, so that the language model can perceive and rely on the known semantic edges in the graph structure during context processing.

Through the above modeling method, the proposed algorithm can achieve deep collaboration between the language model and the knowledge graph at the structural level, effectively enhance the model's ability to represent entity semantics, contextual relationships, and reasoning paths, and provide basic support for language understanding and generation in knowledge-intensive tasks.

III. DATASET

This study uses the T-REx (Textual Relations Extraction) knowledge graph dataset as the primary source of structured knowledge. T-REx is a large-scale collection of knowledge triples built from Wikipedia. It contains millions of fact triples in the form of entity-relation-entity, covering a wide range of general-domain concepts. Each triple is aligned with multiple natural language sentences, which express the fact either explicitly or implicitly. This alignment provides valuable support for linking knowledge with language.

The dataset offers high-quality entity annotations and well-organized semantic relations. Its structured representation includes rich hypernym – hyponym links, functional relations, and synonym mappings. These characteristics make it suitable for graph neural network modeling. T-REx also provides aligned sentence pairs for fine-tuning language models. These pairs help integrate contextual representations of entities with the structure of the graph, supporting the learning of factual expressions and logical connections from text.

To improve the relevance of knowledge injection and the adaptability to specific tasks, this study filters domain-specific subsets and prunes the graph structure based on T-REx. Semantic-dense and structurally clear subgraphs are preserved. Entity subgraphs are then constructed for graph neural network modeling. This processing enables the model to receive structured knowledge enhancement signals while preserving semantic completeness. As a result, the model improves its ability to capture semantic dependencies.

IV. PERFORMANCE EVALUATION

This paper first conducts a comparative experiment, and the experimental results are shown in Table 1.

Table1. Comparative experimental results

| Model | QA-Acc | F1-Score | BLEU |
|---|---|---|---|
| KGLM[24] | 78.6% | 74.2% | 21.5% |
| DRAGON[25] | 81.3% | 76.8% | 24.1% |
| KG-SFT[26] | 83.7% | 78.9% | 26.5% |
| **Ours** | 86.4% | 82.1% | 29.7% |

As shown in Table 1, the proposed method achieves consistent improvements across key metrics with the integration of knowledge graph injection. The QA-Acc reaches 86.4%, nearly eight points higher than KGLM, confirming the benefit of explicit knowledge modeling for entity relations and semantic alignment. The method also yields an F1-Score of 82.1%, outperforming models such as DRAGON and KG-SFT, which indicates stronger robustness in entity discrimination and relation reconstruction. In addition, the BLEU score rises to

29.7%, evidencing improved content consistency and contextual coherence through structured knowledge guidance. These results collectively show that knowledge graph-based fine-tuning enhances factual reasoning and semantic reliability, thereby mitigating limitations of pretrained models in knowledge-intensive tasks. The influence of different learning rate settings on performance is further illustrated in Figure 2.

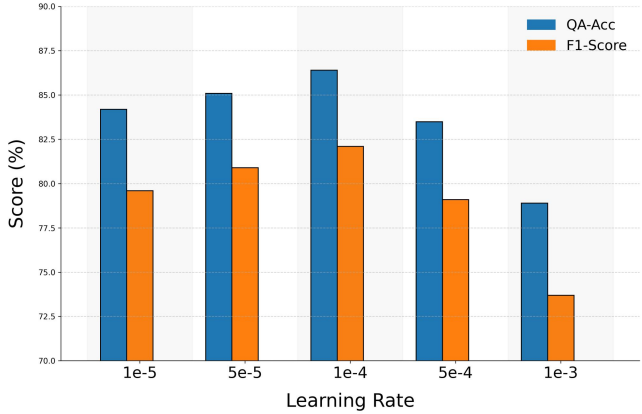

Figure 2. The impact of different learning rate settings on model performance

As shown in Figure 2, the performance of the proposed model is significantly affected by different learning rate settings. Under smaller learning rates such as 1e-5 and 5e-5, the model shows stable and high performance on both QA-Acc and F1-Score, maintaining around 85% and 80% respectively. This indicates that slower parameter updates help achieve stable integration of structured knowledge. It also avoids fluctuations during the injection process, highlighting the sensitivity of knowledge-aware fine-tuning to learning rate settings.

When the learning rate increases to 1e-4, the model reaches peak performance across all metrics. This suggests that the model achieves an optimal balance between semantic modeling and structural guidance. The result indicates that in the context of graph-based knowledge injection, a moderate learning rate helps the language model effectively integrate knowledge graph information while maintaining coherence and accuracy in language modeling.

However, when the learning rate is further increased to 5e-4 and 1e-3, the model's performance drops significantly. F1 scores decline rapidly. This may be related to instability in the structured knowledge injection process. Faster gradient updates hinder the fusion module from learning long-range dependencies from the knowledge graph. As a result, the model loses logical consistency and entity alignment in generation and understanding tasks.

Overall, the experimental results suggest that the structured and controllable knowledge injection mechanism is highly sensitive to the learning rate. A well-chosen learning rate can promote collaborative modeling between the language model and the graph semantics. It can also improve the model's robustness and generalization in complex language understanding tasks. This provides important insights for building structure-aware large models.

This paper also gives the impact of graph subgraph coverage on entity prediction accuracy, and the experimental results are shown in Figure 3.

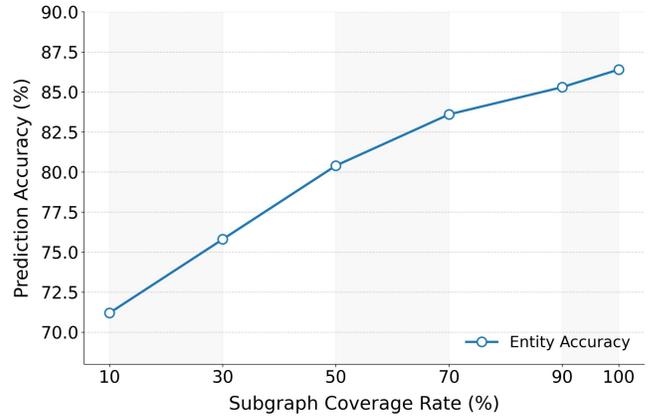

Figure 3. The impact of graph subgraph coverage on entity prediction accuracy

As shown in Figure 3, entity prediction accuracy steadily increases as the subgraph coverage rate rises, showing a clear positive correlation. When the subgraph coverage is low, such as 10% or 30%, the model's ability to recognize entities is limited. The prediction accuracy remains between 71.2% and 75.8%. This suggests that with insufficient structural knowledge, the language model struggles to capture entity relations and semantic dependencies in context.

When the coverage rate reaches 50% to 70%, the accuracy improves rapidly to over 80%. This indicates that enhanced graph structure information significantly improves semantic mapping and logical modeling between entities. In this coverage range, the model gains more hypernyms, synonyms, and functional relation data. The graph context begins to play a compensatory role, enhancing the completeness of reasoning paths and the ability to distinguish entity boundaries.

As the coverage increases to 90% and above, the accuracy gain becomes marginal, approaching a ceiling of 86.4%. This convergence effect suggests that at high coverage levels, the graph provides nearly complete structural context. The model no longer relies on redundant knowledge and consistently generates high-quality predictions. This reflects the saturation point of structural guidance in improving representation quality.

V. CONCLUSION

This paper addresses the challenges faced by large language models in knowledge-intensive tasks, particularly the lack of structural information and weak reasoning capability. It proposes a fine-tuning method based on knowledge graph injection. By introducing structured knowledge, the method enhances pretrained models in a targeted manner, improving knowledge completion and logical reasoning during language understanding, entity recognition, and contextual modeling. The architecture integrates a graph neural network to encode the knowledge graph and fuses it with the hidden representations of the language model. This builds a knowledge-aware semantic representation path that promotes structure-controllable and semantically consistent outputs in language generation and reasoning.

Experimental results show that the proposed method achieves superior performance across multiple key metrics. It performs particularly well in entity prediction, question answering accuracy, and semantic generation quality. The model demonstrates strong adaptability and stability concerning knowledge injection. Systematic sensitivity experiments further validate how the model responds to knowledge sparsity, graph structure perturbations, and hyperparameter changes. These findings provide quantifiable guidance for model deployment and tuning. The results indicate that incorporating structured knowledge during fine-tuning not only enhances the expression of explicit knowledge but also pushes semantic modeling toward structure-aware mechanisms.

In broader application contexts, the proposed fine-tuning paradigm holds significant value for domains such as financial analysis, medical document processing, and legal text understanding. By integrating entities, relations, and logical chains from knowledge graphs, language models gain the ability to decompose, generalize, and reason over complex sentences. This enhances the interpretability, controllability, and generalization of models in specialized and high-risk domains. It also addresses the problem of discontinuous and inconsistent knowledge arising from memory-based reasoning in traditional models, offering structural support for intelligent systems in complex tasks.

Future research may explore mechanisms for dynamically injecting knowledge graphs, allowing language models to adaptively select the most relevant subgraphs based on task context. Another important direction is to build structure-aware fusion paths across modalities and achieve efficient knowledge transfer in low-resource domains. Overall, the structure-injection fine-tuning strategy presented in this paper provides a foundation for building trustworthy, explainable, and domain-proficient language models. It also opens new perspectives for developing knowledge-enhanced intelligent language systems.

## REFERENCES

[1] T. Susnjak, P. Hwang, N. Reyes et al., "Automating research synthesis with domain-specific large language model fine-tuning," ACM Transactions on Knowledge Discovery from Data, vol. 19, no. 3, pp. 1–39, 2025.

[2] R. Pan, X. Liu, S. Diao et al., "Lisa: Layerwise importance sampling for memory-efficient large language model fine-tuning," Advances in Neural Information Processing Systems, vol. 37, pp. 57018–57049, 2024.

[3] T. T. Kim, M. Makutonin, R. Sirous et al., "Optimizing large language models in radiology and mitigating pitfalls: Prompt engineering and fine-tuning," RadioGraphics, vol. 45, no. 4, Art. no. e240073, 2025.

[4] W. Lu, R. K. Luu, and M. J. Buehler, "Fine-tuning large language models for domain adaptation: Exploration of training strategies, scaling, model merging and synergistic capabilities," npj Computational Materials, vol. 11, no. 1, Art. no. 84, 2025.

[5] J. Liu, X. Gu, H. Feng, Z. Yang, Q. Bao, and Z. Xu, "Market turbulence prediction and risk control with improved A3C reinforcement learning," Proceedings of the 2025 8th International Conference on Advanced Algorithms and Control Engineering (ICAACE), pp. 2634-2638, 2025.

[6] X. Li, Y. Peng, X. Sun, Y. Duan, Z. Fang, and T. Tang, "Unsupervised detection of fraudulent transactions in e-commerce using contrastive learning," Proceedings of the 2025 4th International Symposium on Computer Applications and Information Technology, pp. 1663-1667, 2025.

[7] X. Zhang and X. Wang, "Domain-adaptive organ segmentation through SegFormer architecture in clinical imaging," Transactions on Computational and Scientific Methods, vol. 5, no. 7, 2025.

[8] Y. Zi and X. Deng, "Joint modeling of medical images and clinical text for early diabetes risk detection," Journal of Computer Technology and Software, vol. 4, no. 7, 2025.

[9] B. Fang and D. Gao, "Collaborative multi-agent reinforcement learning approach for elastic cloud resource scaling," arXiv:2507.00550, 2025.

[10] X. Sun, Y. Duan, Y. Deng, F. Guo, G. Cai, and Y. Peng, "Dynamic operating system scheduling using double DQN: A reinforcement learning approach to task optimization," Proceedings of the 2025 8th International Conference on Advanced Algorithms and Control Engineering (ICAACE), pp. 1492-1497, 2025.

[11] B. Weng, "Navigating the landscape of large language models: A comprehensive review and analysis of paradigms and fine-tuning strategies," arXiv:2404.09022, 2024.

[12] S. Chaudhary, U. Dinesha, D. Kalathil et al., "Risk-averse fine-tuning of large language models," Advances in Neural Information Processing Systems, vol. 37, pp. 107003–107038, 2024.

[13] W. Zhang, Q. Wang, X. Kong et al., "Fine-tuning large language models for chemical text mining," Chemical Science, vol. 15, no. 27, pp. 10600–10611, 2024.

[14] E. Vulpescu and M. Beldean, "Optimized fine-tuning of large language model for better topic categorization with limited data," 2024.

[15] S. Lyu, Y. Deng, G. Liu, Z. Qi, and R. Wang, "Transferable modeling strategies for low-resource LLM tasks: A prompt and alignment-based," arXiv:2507.00601, 2025.

[16] Y. Wang, "Gradient-guided adversarial sample construction for robustness evaluation in language model inference," Transactions on Computational and Scientific Methods, vol. 4, no. 7, 2024.

[17] Y. Sun, R. Zhang, R. Meng, L. Lian, H. Wang, and X. Quan, "Fusion-based retrieval-augmented generation for complex question answering with LLMs," 2025.

[18] T. Xu, X. Deng, X. Meng, H. Yang, and Y. Wu, "Clinical NLP with attention-based deep learning for multi-disease prediction," arXiv:2507.01437, 2025.

[19] R. Meng, H. Wang, Y. Sun, Q. Wu, L. Lian, and R. Zhang, "Behavioral anomaly detection in distributed systems via federated contrastive learning," arXiv:2506.19246, 2025.

[20] W. Zhu, "Fast adaptation pipeline for LLMs through structured gradient approximation," Journal of Computer Technology and Software, vol. 3, no. 6, 2024.

[21] Y. Qin, "Hierarchical semantic-structural encoding for compliance risk detection with LLMs," Transactions on Computational and Scientific Methods, vol. 4, no. 6, 2024.

[22] Q. Wu, "Task-aware structural reconfiguration for parameter-efficient fine-tuning of LLMs," Journal of Computer Technology and Software, vol. 3, no. 6, 2024.

[23] Y. Zhang, J. Liu, J. Wang, L. Dai, F. Guo, and G. Cai, "Federated learning for cross-domain data privacy: A distributed approach to secure collaboration," arXiv:2504.00282, 2025.

[24] J. Youn and I. Tagkopoulos, "Kglm: Integrating knowledge graph structure in language models for link prediction," arXiv:2211.02744, 2022.

[25] M. Yasunaga, A. Bosselut, H. Ren et al., "Deep bidirectional language-knowledge graph pretraining," Advances in Neural Information Processing Systems, vol. 35, pp. 37309–37323, 2022.

[26] H. Chen, X. Shen, J. Wang et al., "Knowledge graph finetuning enhances knowledge manipulation in large language models," Proceedings of the Thirteenth International Conference on Learning Representations, 2025.